\RequirePackage[svgnames]{xcolor}
\documentclass[11pt,letterpaper]{mystyle}
\usepackage[numbers,sort&compress]{natbib}
\usepackage{xspace}
\usepackage{iftex}
\ifXeTeX\microtypesetup{tracking=false}\fi

\usepackage{amsmath,amsfonts,bm}

\def\eqref#1{equation~\ref{#1}}

\def\1{\bm{1}}

\DeclareMathAlphabet{\mathsfit}{\encodingdefault}{\sfdefault}{m}{sl}
\SetMathAlphabet{\mathsfit}{bold}{\encodingdefault}{\sfdefault}{bx}{n}

\usepackage{amsmath}
\usepackage{amssymb}
\usepackage{amsfonts}
\usepackage{booktabs}
\usepackage{microtype}
\usepackage{graphicx}
\usepackage{multirow}
\usepackage{array}
\usepackage{tabularx}
\usepackage{xspace}
\usepackage{pifont}
\usepackage{xcolor}
\usepackage{colortbl}
\tcbuselibrary{most}
\usepackage{tikz}
\usetikzlibrary{positioning}
\usepackage{placeins}
\usepackage{float}
\usepackage{wrapfig}
\usepackage{hyperref}
\usepackage{url}
\usepackage{fontawesome5}

\hypersetup{colorlinks=true,allcolors=TinaCrimson}

\newcommand{\solveedit}{\textsc{SolveEdit}\xspace}
\newcommand{\solvescore}{\textsc{SolveScore}\xspace}
\newcommand{\solveeditplan}{\textsc{SolveEdit-Plan}\xspace}

\newcommand{\genericvisionrewrite}{\textsc{Generic Vision Rewrite}\xspace}

\newcommand{\markone}{\ding{182}}
\newcommand{\marktwo}{\ding{183}}
\newcommand{\markthree}{\ding{184}}

\newcommand{\result}[1]{#1}

\definecolor{solvebest}{RGB}{218,239,221}
\definecolor{solvesecond}{RGB}{255,238,190}
\definecolor{solvequestion}{RGB}{245,252,253}

\title{\textsc{SolveEdit}: Benchmarking Visual Problem Solving in Generative Models}

\author[1,\textdagger]{Wenjie~Shu}
\author[2,\textdagger]{Yexin~Liu}
\author[2]{Harold~Haodong~Chen}
\author[3]{Xuerui~Qiu}
\author[4]{Zehan~Wang}
\author[1]{Yidi~Zhang}
\author[1]{Yizhan~Chen}
\author[1]{Zunwei~Wang}
\author[5]{Minghao~Liu}
\author[1,*]{Qi~Chen}
\author[2,*]{Harry~Yang}
\author[4]{Xiaogang~Xu}
\affil[1]{ZODA}
\affil[2]{HKUST}
\affil[3]{UCAS}
\affil[4]{ZJU}
\affil[5]{UTokyo\protect\\
  \textsuperscript{\textdagger}Equal contribution.\quad
  \textsuperscript{*}Corresponding authors.}

\runningtitle{SolveEdit: Benchmarking Visual Problem Solving in Generative Models}

\begin{document}

\begin{abstract}
Machine intelligence is often evaluated through abstract reasoning problems, yet many real-world problems are visual, such as arranging objects, repairing layouts, or tracing routes. Solving these problems requires understanding a scene, inferring what must change to achieve a goal, and realizing that change without disturbing unrelated content. However, existing benchmarks mainly evaluate perception, generation, or explicitly specified transformations, leaving goal-driven visual problem solving underexplored. To bridge this gap, we introduce \solveedit, a benchmark for visual problem solving through scene transformation. Given an image and a goal, a model must infer a valid transformation from the request, the scene, or a visually expressed rule, then execute it while preserving unrelated content. \solveedit contains 2,728 cases. Atomic transition contracts specify required and protected conditions, enabling \solvescore to measure completion and unintended changes without a single reference output. The strongest evaluated model achieves only 57.0\% \solvescore. We further introduce \solveeditplan, a two-stage visual planner that instantiates the transition before generation. Under matched single-generation evaluation, it improves \solvescore by 9.1 points on average across three tested generators, including a gain from 57.0\% to 71.6\% for GPT-Image-2, without modifying the editor.

\par\vspace{0.4em}
\noindent{\small
\faGlobe\ \textbf{Project}\quad\url{https://wenjieshu.github.io/SolveEdit-project-page/}\par
\faGithub\ \textbf{Code}\quad\url{https://github.com/WenjieShu/SolveEdit}}

\end{abstract}

\maketitle

\section{Introduction}
\label{sec:intro}
\vspace{-0.4em}

\begin{wrapfigure}{r}{0.54\textwidth}
\vspace{-2.4em}
 \centering
 \includegraphics[width=\linewidth]{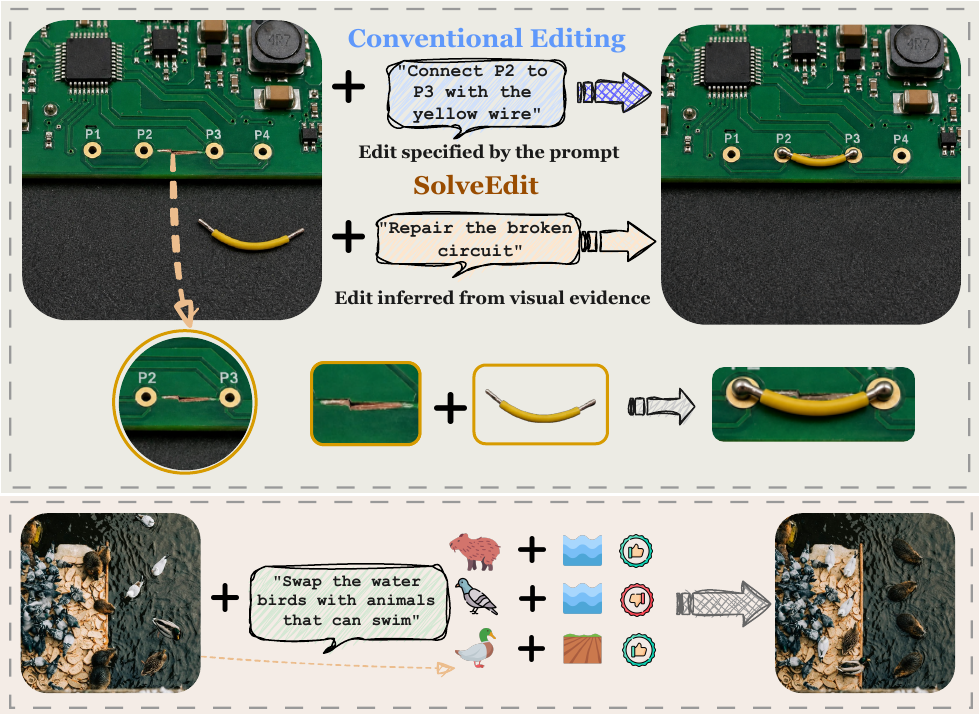}
 \vspace{-1.8em}
 \caption{\solveedit: infer, execute, preserve.}
 \vspace{-1.6em}
 \label{fig:teaser}
\end{wrapfigure}

Machine intelligence is usually tested with abstract reasoning problems,
but much real-world problem solving starts in front of a visual scene, as in
Figure~\ref{fig:teaser}. Given the scene and a goal, the system has to
understand the current state, work out what should change, produce a valid
final state, and leave unrelated content alone. Visual interpretation,
transition determination, and execution all enter the process. We study this
capability as \emph{visual problem solving through scene transformation}:
the model answers by changing the scene, not by returning text.

Current evaluations tend to isolate perception, abstract reasoning,
generation, or the execution of explicitly specified transformations.
Instruction-based editing benchmarks, for example, check whether a model can
carry out a stated change while preserving the rest of an
image~\citep{brooks2023instructpix2pix,zhang2023magicbrush,wang2023editbench,ku2024imagenhub,ma2024i2ebench}.
Newer image-to-image benchmarks add temporal, causal, spatial, logical,
factual, procedural, and planning tasks~\citep{zhao2025risebench,wu2025kris,han2026unireditbench}.
What they do not isolate is where the required transformation comes from:
the request, the current scene, or a rule expressed in the image. Executing
a specified transformation and inferring an unresolved one can therefore be
scored together. Reference-based scoring adds a second problem: valid
solutions that differ from the single authored output may be penalized. These
gaps lead to our central question: \emph{Can a generative model determine a
valid transformation from a source image and a goal, and then realize it in
the same scene?}


We answer with \solveedit, which puts transition determination inside the
visual problem instead of supplying it through the prompt. The distinction
matters. A model can execute a familiar edit pattern without working out
what the current scene requires, just as a language model that has memorized
facts or solution templates need not generalize to a new configuration. We
therefore control the source of the information that fixes a valid
transition. The 2,728 cases, spread over 10 domains and 54 subdomains, fall
into three regimes. In \textbf{Instruction-Specified (IS)} cases the prompt
specifies the change. In \textbf{State-Dependent (SD)} cases the model
resolves a missing variable from observable scene state or basic spatial and
structural relations. In \textbf{Rule-Dependent (RD)} cases it interprets a
rule expressed in the image, sometimes with knowledge beyond the immediate
scene. Removing a grounded marker is IS; placing a block in the only empty
slot is SD; sorting pieces by a depicted legend is RD. Instruction
following, scene understanding, and rule inference are thus separated within
one generative task. The axis is orthogonal to content domain and task
type. The same spatial operation can land in IS, SD, or RD depending on what
fixes its valid outcome.

More than one solution can be valid, so each case carries an atomic
\emph{transition contract}. Required conditions state what the output must
accomplish; protected conditions state what must remain intact. \solvescore
measures required completion, applies a bounded deduction for unintended
changes, and reports six diagnostic scores across the required and protected
semantic, relational, and visual-quality conditions. The same contracts
score image edits and final frames from video generators.


We evaluate nine image-to-image models and two image-to-video models on the
same contracts. The strongest of them reaches only
\result{57.0\%} \solvescore, which leaves substantial room for improvement.
Its diagnostic scores show larger deficits in required semantic and
relational conditions than in visual-quality ones; solution recovery appears
to be a large part of the failure.
That diagnosis suggests a direct test: determine the transition before
generation. We run it with \solveeditplan, a
two-stage agentic visual planner that instantiates scene-dependent transition
variables and writes an evidence-grounded editing instruction, with no
training or changes to the editor. Under matched single-generation evaluation,
it takes GPT-Image-2 from \result{57.0\%} to \result{71.6\%} \solvescore,
transfers to the open-source editor and the video generator, and beats a
call-matched \genericvisionrewrite control by \result{6.3} points on
GPT-Image-2, with higher required completion and less collateral damage overall.

Our contributions are threefold:
\begin{itemize}
    \setlength{\topsep}{0pt}
    \setlength{\partopsep}{0pt}
    \setlength{\itemsep}{0.3em}
    \setlength{\parsep}{0pt}
    \setlength{\parskip}{0pt}
    \item[\markone] \textbf{A problem formulation and benchmark.} We formulate visual problem solving through scene transformation and introduce \solveedit, a benchmark of 2,728 cases across 10 application domains and 54 subdomains, organized by the information source that determines the required transition.
    \item[\marktwo] \textbf{A contract-based evaluation framework.} We define atomic transition contracts and \solvescore to accommodate multiple valid visual outcomes, separate required completion from collateral changes, and give six diagnostic views of model behavior.
    \item[\markthree] \textbf{A diagnosis and targeted intervention.} We analyze current image and video generators across the IS, SD, and RD regimes, and introduce \solveeditplan to test whether explicitly instantiating the transition addresses the observed deficits in required semantic and relational conditions.
\end{itemize}

\section{Related Work}
\label{sec:related}
\vspace{-0.4em}

\begin{figure}[!t]
\centering
\includegraphics[width=\textwidth]{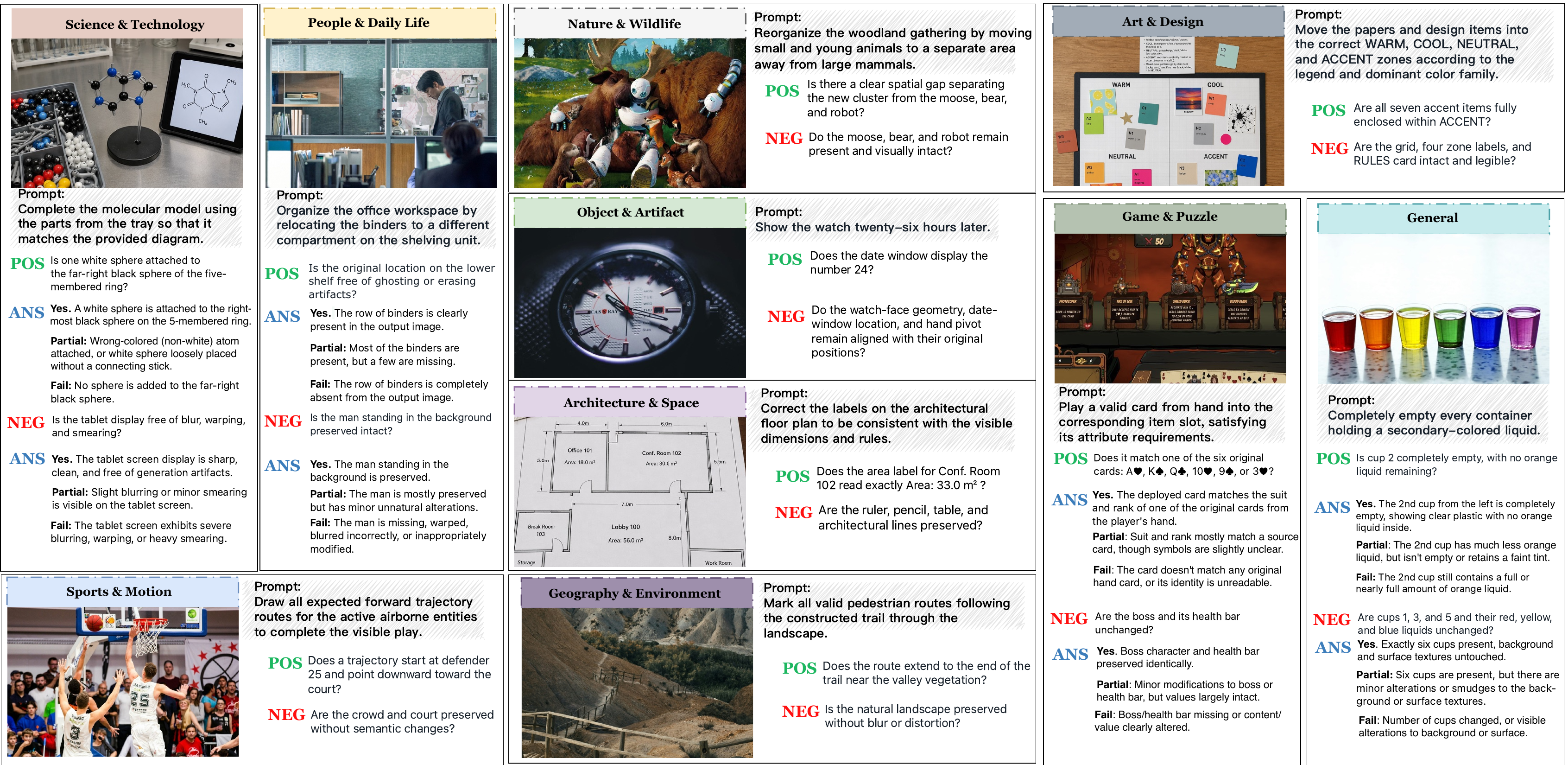}
\caption{\textbf{Application breadth of \solveedit.} One representative case from each of ten domains. Panels show the source, instruction, evaluator questions, and reference answers.}
\label{fig:domain_coverage}
\end{figure}

\paragraph{Visual Reasoning and Image Editing.}
Visual reasoning has progressed from language or discrete-answer benchmarks to
tasks that infer transformations or realize answers as images and videos~\citep{goyal2017vqa,johnson2017clevr,hudson2019gqa,suhr2019nlvr2,marino2019okvqa,yue2024mmmu,chollet2019measure,chen2026babyvision,li2026corebench,chen2025tivibench}. Image-editing methods generally receive the desired change explicitly, including both instruction-based~\citep{brooks2023instructpix2pix,zhang2023magicbrush,sheynin2024emuedit,hui2024hqedit,zhao2024ultraedit,yu2025anyedit} and training-free approaches~\citep{meng2021sdedit,hertz2022prompttoprompt,mokady2023nulltext,kawar2023imagic,cao2023masactrl,parmar2023pix2pixzero,tumanyan2023pnpdiffusion,brack2024leditspp}. More recent benchmarks extend editing evaluation to grounded correctness, preservation, knowledge, reasoning, and planning~\citep{wang2023editbench,ku2024imagenhub,ma2024i2ebench,qian2025giebench,zhao2025risebench,wu2025kris,han2026unireditbench,zhang2026vibe,zheng2026uniedit}. Our focus is complementary: \solveedit makes the source of transition-defining information an explicit evaluation axis and uses required and protected contracts to allow multiple valid renderings while measuring collateral changes. It also spans heterogeneous visual sources, including photographs, animation, games, cinematic frames, and purpose-built scenes, testing whether transition recovery transfers across visual styles. Figure~\ref{fig:domain_coverage} shows this breadth.

\vspace{-0.8em}
\paragraph{Evaluation of Generated Visual Content.}
Distributional and embedding metrics measure fidelity or cross-modal alignment~\citep{heusel2017fid,radford2021clip,hessel2021clipscore}; learned preference and object-focused evaluators provide more targeted text-to-image signals~\citep{xu2023imagereward,wu2023hps,ghosh2023geneval}; and question-based evaluators decompose outputs into interpretable judgments~\citep{hu2023tifa,ku2024viescore,cho2024dsg,wu2023qbench}. These metrics primarily score an image or its alignment with text. In contrast, \solveedit evaluates a transition between an input and an output. Its atomic contracts separate required completion from independently protected content, allow multiple valid final states, and report collateral damage separately from task completion.

\vspace{-0.8em}
\paragraph{Multimodal Reasoning for Visual Generation.}
Reasoning-action agents, tool-using models, and multimodal language models provide foundations for visually grounded planning~\citep{yao2023react,schick2023toolformer,driess2023palme,wang2024qwen2vl,wang2024internvl}. Visual agents compose tools or foundation models for multi-step generation and editing~\citep{yang2023mmreact,wu2023visualchatgpt,wang2024genartist}, while editing systems increasingly use multimodal understanding, intent interpretation, or intermediate plans~\citep{fu2024mgie,huang2024smartedit,hu2024instructimagen,ji2025planningediting,ci2025describedontdictate}. SolveEdit-Plan addresses a narrower question: whether a task-specific transition can be recovered before one call to an unchanged generator. It uses targeted observation, candidate comparison, and explicit preservation obligations, with a matched planning backend, call budget, and final generator. For video, we use terminal frames only to test transfer of the same transition contracts across modalities; temporal generation is outside our scope~\citep{kondratyuk2023videopoet,bartal2024lumiere,yang2024cogvideox,kong2024hunyuanvideo,huang2024vbench,fu2025videomme}.

\section{\solveedit: Formulation, Benchmark, and Evaluation}
\label{sec:problem}
\label{sec:benchmark}
\vspace{-0.4em}

\begin{figure}[t]
\centering
\includegraphics[width=\textwidth]{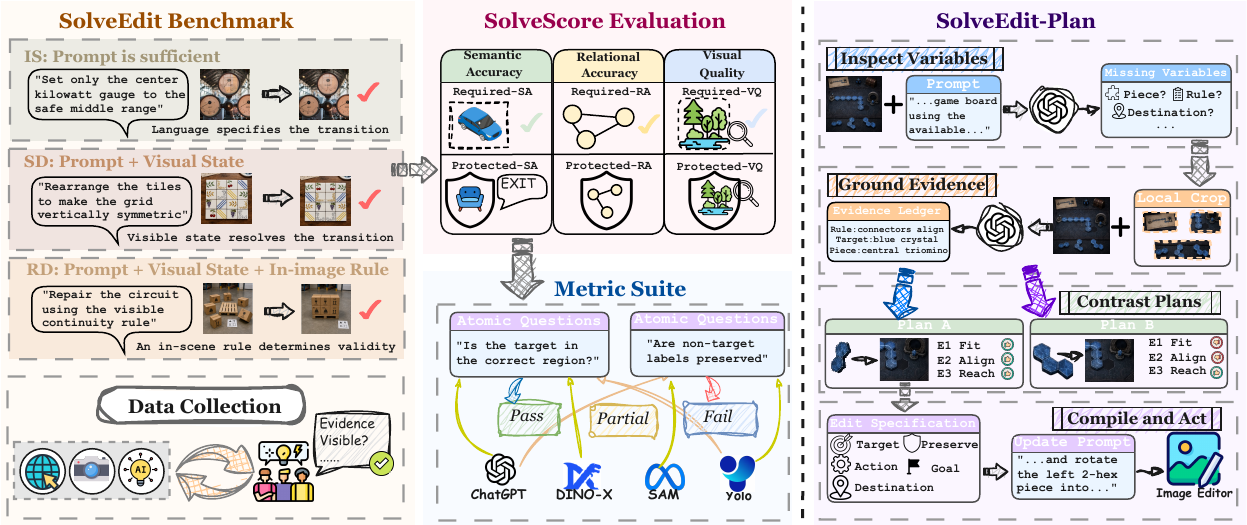}
\vspace{-1.6em}
\caption{\textbf{Overview of \solveedit, \solvescore, and \solveeditplan.}
\textbf{Left:} cases are grouped by the information needed to determine a valid edit.
\textbf{Center:} atomic criteria specify the required change and the visual state to preserve; \solvescore aggregates their satisfaction.
\textbf{Right:} \solveeditplan instantiates the transition and compiles an instruction for an unmodified editor.}
\vspace{-1.2em}
\label{fig:solveedit_overview}
\end{figure}

\vspace{-0.4em}
\subsection{Task Formulation and Dependency Regimes}
\vspace{-0.4em}

Let $I$ be an input image and $x$ a natural-language request. Writing the
editor as $Y=f(I,x)$ hides the edit that satisfies $x$ in this particular
scene. We make that edit explicit as a specification $z$ listing the relevant
entities, the action, the final state, and the preservation scope. Correctness is not one target image. Several specifications $z$ can
satisfy the same request, and we collect them in the set
$\mathcal{Z}^{*}(I,x)$. Figure~\ref{fig:solveedit_overview} places this set
next to the transition contract and the planning interface.

Once visible entities are grounded, the remaining question is what fixes
$z$. The image is always used for localization and rendering, so it stays in
the pipeline regardless of regime. Four ingredients can enter:
request-to-entity bindings $B(I,x)$, background knowledge $K$, current-scene
facts $S(I)$, and an in-scene rule $\mathcal{R}(I)$ (a legend, a constraint,
or a repeated pattern). Eq.~\eqref{eq:solution_sources} writes the admissible
set as a function $\Phi$ of these four:
\begin{equation}
    \mathcal{Z}^{*}(I,x)=\Phi\!\left(x,B(I,x),K,S(I),\mathcal{R}(I)\right).
    \label{eq:solution_sources}
\end{equation}
A generated image $Y$ belongs to $\mathcal{Y}^{*}(I,x)$ under two conditions:
it realizes some $z\in\mathcal{Z}^{*}(I,x)$, and visual state outside that
$z$ is preserved:
\begin{equation}
    \mathcal{Y}^{*}(I,x)=\left\{Y:\exists z\in\mathcal{Z}^{*}(I,x),\;
    \operatorname{Realize}(Y;I,z)\land\operatorname{Preserve}(Y;I,z)\right\}.
    \label{eq:valid_outputs}
\end{equation}
Two renderings of the same $z$ both count. A plausible rendering of the
wrong $z$ does not.


The three regimes do not overlap, even though background knowledge $K$ may
be used in any of them. The label is assigned after grounding and $K$ are
applied: it records which case-specific piece of information fixes the
solution. Under that rule, \textbf{IS} means the grounded request already
fixes $z$; \textbf{SD} means a current-scene fact resolves a variable the
request leaves open (target, action, destination, route, or final state);
\textbf{RD} means the model also reads a rule shown or instantiated in the
image. The label uses the minimum information needed, so it is not a
difficulty rating. Appendix~\ref{app:annotation} gives the decision
protocol, including examples that distinguish adjacent regimes.

Annotators follow a fixed procedure. They first bind the request to visible
entities, then ask whether the grounded request together with $K$ fixes the
transition. If it does not, a current-scene fact or a case-local rule must
resolve the remaining variable, and that fact decides the label. Ordinary
localization stays in IS; knowledge that never appears in the image cannot
make a case RD. The label is therefore operational, not a claim about model
architecture: IS cases still require visual grounding, and an RD case can be
easier than an IS case under the same protocol.

A case can fail in two distinct ways. The model may settle on an invalid
$z$, choosing an occupied destination or applying the wrong visible rule. Or
it may recover a valid $z$ and fail to render it: the route comes back
disconnected, an object lands in the wrong place, or unrelated content
changes. We call the first a solution-recovery error and the second a
visual-execution error. The evaluation covers both: it scores the solved
state and, separately, the content that should have stayed unchanged.

\vspace{-0.4em}
\subsection{Benchmark Construction and Coverage}
\vspace{-0.4em}

The 2,728 cases are each built around a visual problem in a concrete
application. The request given to the model states only the goal. It never
reveals the scene state or rule attached to the case's regime. Behind each
case, the authors record the evidence needed to resolve the request, mark one
or more admissible final states, and mark the content allowed to change. A
case survives review only when four conditions hold: the intended change is
semantically checkable; its evidence is visible in the image; correctness can
be stated without matching an exact reference; and the change can be separated
from protected content. Cases that fail, or rest on private assumptions, are
revised or rejected before entering the manifest.

Each retained case is stored as
\begin{equation}
    c=\left(I,x,m,\mathcal{C}^{\mathrm{req}},\mathcal{C}^{\mathrm{pro}},\mathcal{A}\right),
    \label{eq:case_representation}
\end{equation}
Here $m$ holds descriptive metadata, $\mathcal{C}^{\mathrm{req}}$ and
$\mathcal{C}^{\mathrm{pro}}$ hold the required and protected conditions, and
$\mathcal{A}$ holds optional audit assets. A reference edit may be attached
to show one admissible solution; it has no role in defining correctness. The
request shown to the model and the evidence used to annotate the case are
stored separately from the model input.

\begin{figure*}[t]
\centering
\includegraphics[width=\textwidth]{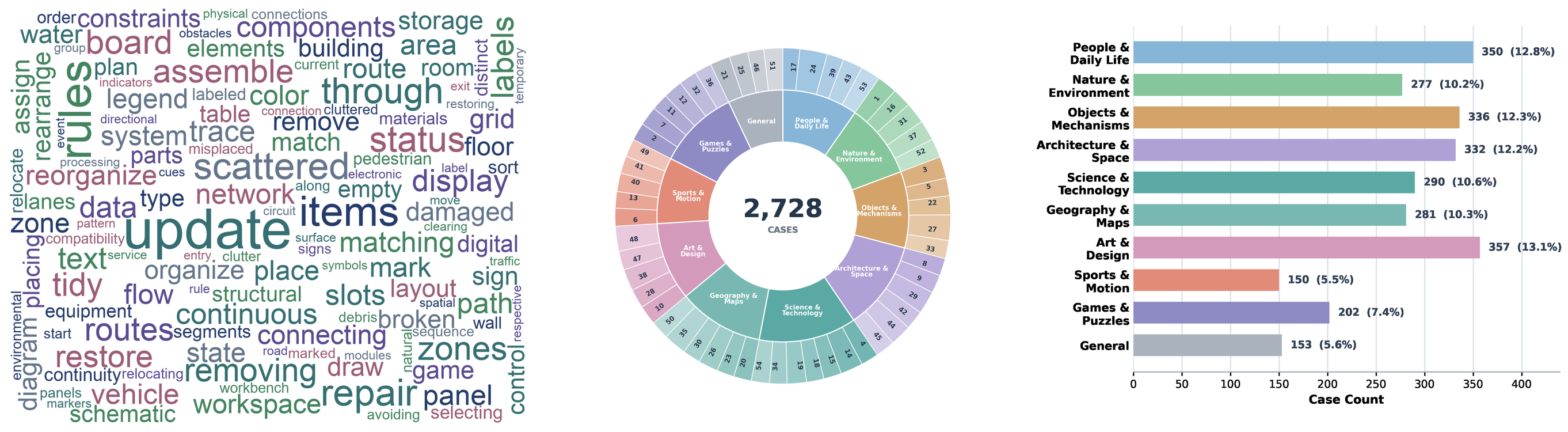}
\vspace{-1.8em}
\caption{Statistics of the 2,728-case \solveedit benchmark. \textbf{Left:} frequent request terms. \textbf{Center:} coverage across 10 domains and 54 subdomains. \textbf{Right:} cases per domain.}
\vspace{-1.5em}
\label{fig:benchmark_statistics}
\end{figure*}

The 10 domains and 54 subdomains cover people and everyday life, natural
and environmental systems, objects and mechanisms, architecture and space,
science and technology, geography and maps, art and design, sports and
motion, games and puzzles, and general scenes. A domain label records the
setting rather than the entities, and it is independent of the regime.
Photographic, animated, cinematic, game, and purpose-built imagery all
appear, with underrepresented settings and transitions authored on purpose.
Figure~\ref{fig:benchmark_statistics} gives this coverage and the request
distribution. Source diversity alone does not qualify a case: the image has
to support a well-defined problem that can be checked visually.

Quality control checks image validity, request grounding, contract coverage,
dependency evidence, and split integrity. Shared scenes and near duplicates
are grouped before splitting. Cases are revised or withheld if the request,
visible evidence, and hidden contract disagree. Appendix~\ref{app:dataset}
gives the review procedure.

\vspace{-0.4em}
\subsection{Atomic Evaluation and \solvescore}
\vspace{-0.4em}

No one reference image covers $\mathcal{Y}^{*}(I,x)$. We therefore pair
each case with an atomic \emph{transition contract}. Every criterion $q$
makes a single observable claim, with the pass, partial, and fail conditions
and the evidence needed to judge it written out. The evaluator returns
$a_q\in\{1,0.5,0\}$. It abstains when the input or output lacks sufficient visible evidence.

Criteria have two roles and three diagnostic properties. Required criteria
state what the edit must accomplish, and protected criteria state what must
remain intact. A target car left outside a valid space fails a required
condition; a second car moved by mistake fails an independent protected one.
We keep a protected criterion only when it can fail with every required
condition passing, so one error is never counted twice.

SA covers entities, identities, counts, attributes, text, symbols, and
intrinsic states. RA covers spatial and structural relations: placement,
ordering, containment, connectivity, topology. VQ covers visible rendering
defects such as residual traces, malformed geometry, and illegible content.
Crossing the two roles with the three properties gives six cells. For role
$r\in\{\mathrm{req},\mathrm{pro}\}$ and property
$k\in\{\mathrm{SA},\mathrm{RA},\mathrm{VQ}\}$:
\begin{equation}
    S_{r,k}=
    \frac{\sum_{q:r(q)=r,\,d(q)=k,\,a_q\neq\bot}\omega_q a_q}
         {\sum_{q:r(q)=r,\,d(q)=k,\,a_q\neq\bot}\omega_q},
    \label{eq:dimension_score}
\end{equation}
$a_q=\bot$ marks an abstention, and only non-abstained atoms enter the sums.
All six scores are higher-is-better. The weights are hierarchical, so
redundant criteria or mechanically split atoms cannot inflate a requirement.

With normalized weights $w_q$ and $u_q$, each summing to one over the
required and protected criteria, completion and damage are
\begin{equation}
    R=\sum_{q\in\mathcal{C}^{\mathrm{req}}} w_q a_q,
    \qquad
    D=\sum_{q\in\mathcal{C}^{\mathrm{pro}}} u_q(1-a_q).
    \label{eq:required_damage}
\end{equation}
Abstained criteria add zero without redistributing weight. The case score
is
\begin{equation}
    \solvescore_{\lambda}
    = G_{\mathrm{quality}}\max\!\left(0, R-\lambda D\right),
    \label{eq:solvescore}
\end{equation}
$G_{\mathrm{quality}}$ rejects outputs that are missing, blank, severely
corrupted, unrelated to the request, or unusable. The primary protocol uses
$\lambda=0.5$: completion stays the main objective, and damage can remove at
most half the score. Cases are scored individually before averaging. We
report $R$, $D$, the six diagnostics, the quality-gate pass rate, and
coverage separately; coverage is the share of criterion weight with a
non-abstained verdict. Appendix~\ref{app:contracts} covers weights,
abstentions, and authoring rules.


A VLM performs most of the evaluation. A fixed subset of criteria is also
assigned to specialized checkers beforehand, and the assignment is the same
for every model. These are the conditions where VLM judgments are less
reliable or a tool gives a more objective check. The evaluator sees the
input, output, request, and full contract; the generator never sees the
contract. For assigned criteria, SAM and YOLO supply segmentation and
detection evidence next to the VLM's interpretation. Their output is a
quality gate plus atomic verdicts with evidence, and a deterministic scorer
aggregates them without further judgment. All editors use the same protocol,
VLM version, and scoring code. Generation failures, gate failures, coverage,
and abstentions are reported apart from the aggregate score, making the
limits of the available evaluation evidence visible.

\section{Planning Before Editing}
\label{sec:solveedit_plan}
\vspace{-0.4em}

\solveeditplan tests whether a model can determine the transition before
it renders anything. It is a two-stage test-time planner with no learned
parameters, and the editor is left unchanged. The first stage, Inspect,
finds the unresolved variables and gathers targeted visual evidence. The
second, Resolve, compares plausible transitions and compiles the chosen one
into an instruction for a single call to the original editor:
\begin{equation}
    (\hat{z},\hat{O},\tilde{x})=H(I,x),
    \qquad \hat{Y}=f(I,\tilde{x}),
    \label{eq:solveedit_plan}
\end{equation}
Here $\hat{z}$ is the selected transition, $\hat{O}$ the observable
completion and preservation obligations, and $\tilde{x}$ the instruction
handed to the unchanged final generator.


\paragraph{Transition planning.}
Inspect looks for variables the request leaves unresolved: target, action,
destination, final state, preservation scope. It requests targeted crops for
those variables and records the evidence each crop provides. Resolve then
compares the plausible transitions against that evidence. The transition it
selects is compiled into completion conditions and preservation obligations
for the final editing call to the unchanged generator, including the scene content to preserve.

\vspace{-0.8em}
\paragraph{Generation interface and control.}
The compiled instruction goes to the original generator in one call. The
obligations in $\hat{O}$ shape that instruction, but they are not the
human-authored contract \solvescore uses, and the intermediate transition
gets no supervision or evaluator feedback. At inference the planner sees only
the source image and the goal: no reference outputs, contracts, authored
regions, manual facts, dependency labels, or evaluator responses. The matched
control, \genericvisionrewrite, runs on the same backend with two agent calls
and one final generation. It returns an unconstrained rewritten instruction,
without explicit transition variables, candidate comparison, or preservation
obligations. Appendix~\ref{app:solveedit_plan_results} gives the agent
configuration, schemas, accounting rules, and implementation details.

\section{Experiments}
\label{sec:experiments}
\vspace{-0.4em}

\begingroup
\setlength{\tabcolsep}{0pt}
\renewcommand{\result}[1]{\textcolor{black}{#1}}
\begin{table}[!t]
\centering
\scriptsize
\renewcommand{\arraystretch}{1.12}
\caption{Performance on \solveedit (\%, higher is better). Best \solvescore is bold. Evaluation coverage is reported in Appendix~\ref{app:results}; video outputs are scored from their final frames. SA, RA, and VQ denote Semantic Accuracy, Relational Accuracy, and Visual Quality.}
\label{tab:benchmark_results}
\begin{tabular*}{\textwidth}{@{\extracolsep{\fill}}lccccccccc@{}}
\toprule
\multirow{2}{*}{\textbf{Model}}
& \multicolumn{3}{c}{\textbf{Required}}
& \multicolumn{3}{c}{\textbf{Protected}}
& \multirow{2}{*}{$\mathbf{R}\!\uparrow$}
& \multirow{2}{*}{$\mathbf{D}\!\downarrow$}
& \multirow{2}{*}{\textbf{\solvescore}$\uparrow$} \\
\cmidrule(lr){2-4}\cmidrule(lr){5-7}
& \textbf{SA}$\uparrow$ & \textbf{RA}$\uparrow$ & \textbf{VQ}$\uparrow$
& \textbf{SA}$\uparrow$ & \textbf{RA}$\uparrow$ & \textbf{VQ}$\uparrow$ & & & \\
\midrule
\rowcolor{gray!12}\multicolumn{10}{c}{\textit{Image-to-Video Models}} \\
HunyuanVideo-1.5~\citep{wu2025hunyuanvideo} & \result{16.9} & \result{17.9} & \result{22.5} & \result{42.2} & \result{57.4} & \result{41.9} & \result{17.3} & \result{55.5} & \result{7.4} \\
Kling V3~\citep{kuaishou2026kling3} & \result{37.6} & \result{37.1} & \result{54.7} & \result{64.1} & \result{59.7} & \result{75.5} & \result{39.1} & \result{34.0} & \result{27.5} \\
\midrule
\rowcolor{gray!12}\multicolumn{10}{c}{\textit{Open-Source Models}} \\
OmniGen2~\citep{omnigen2} & \result{19.8} & \result{16.2} & \result{25.4} & \result{49.9} & \result{64.3} & \result{53.4} & \result{18.1} & \result{45.9} & \result{9.7} \\
BAGEL~\citep{deng2025emerging} & \result{27.9} & \result{20.0} & \result{24.2} & \result{51.7} & \result{60.7} & \result{53.6} & \result{22.4} & \result{45.5} & \result{12.1} \\
FLUX.1 Kontext [dev]~\citep{bfl2025fluxkontext} & \result{25.4} & \result{20.5} & \result{40.3} & \result{76.4} & \result{75.8} & \result{80.0} & \result{24.2} & \result{22.5} & \result{18.7} \\
Qwen-Image-Edit-2509~\citep{wu2025qwenimagetechnicalreport} & \result{36.4} & \result{28.2} & \result{45.8} & \result{48.1} & \result{53.9} & \result{56.4} & \result{32.7} & \result{47.5} & \result{20.8} \\
FLUX.2 [dev]~\citep{flux-2-2025} & \result{38.9} & \result{30.6} & \result{51.7} & \result{62.9} & \result{65.3} & \result{77.3} & \result{31.7} & \result{32.3} & \result{22.8} \\
\midrule
\rowcolor{gray!12}\multicolumn{10}{c}{\textit{Commercial Models}} \\
Qwen Image 2.0 Pro~\citep{zhao2026qwen} & \result{53.0} & \result{47.6} & \result{68.0} & \result{65.0} & \result{68.7} & \result{81.6} & \result{50.7} & \result{28.8} & \result{40.6} \\
Gemini 3.1 Flash Image~\citep{google2026gemini31flashimage} & \result{68.3} & \result{64.0} & \result{78.0} & \result{71.5} & \result{72.3} & \result{88.4} & \result{66.0} & \result{23.1} & \result{55.5} \\
Seedream 5.0 Pro~\citep{bytedance2026seedream5pro} & \result{65.1} & \result{63.0} & \result{76.0} & \result{79.4} & \result{75.5} & \result{92.2} & \result{64.3} & \result{17.6} & \result{56.6} \\
GPT-Image-2~\citep{openai2026gptimage2} & \result{69.6} & \result{64.6} & \result{81.7} & \result{71.7} & \result{65.2} & \result{90.7} & \result{67.6} & \result{23.9} & \textbf{\result{57.0}} \\
\bottomrule
\end{tabular*}
\end{table}
\endgroup

We organize our experiments around three questions linking performance, diagnosis, and planning. RQ1: How well do current generative models solve \solveedit tasks? RQ2: What do visually plausible failures reveal about the limitations of current models? RQ3: Does explicitly determining the transition before generation improve task completion and preservation with a fixed generator?

\vspace{-0.4em}
\subsection{Experimental Setup}
\vspace{-0.4em}

\paragraph{Models and inputs.}
We evaluate nine image-to-image models (five open-weight and four commercial)
and two image-to-video models on the same 2,728-case benchmark manifest.
Table~\ref{tab:benchmark_results} lists all evaluated systems. In the direct
evaluation, every model receives the same source image and intent-level request,
and we score one output per case. The planning comparisons use GPT-Image-2,
Qwen-Image-Edit-2509, and HunyuanVideo-1.5 with fixed final generators;
controls and planning budgets are described in RQ3.

\vspace{-0.8em}
\paragraph{Evaluation protocol.}
Outputs are scored with the same hidden transition contracts and deterministic
\solvescore aggregation, using $\lambda=0.5$ throughout. The evaluator receives
the source image, request, model output, and contract; the contract is withheld
from the generator. Scores are computed per case and averaged over the full
manifest, giving each case equal overall weight. Failed generations are retried at most twice; missing or unusable
outputs receive zero scores and remain in the denominator. Video outputs are scored
from their final frame with the same contract; this evaluates the completed
visual state and treats video results as a complementary cross-modal study,
not a temporal-reasoning benchmark. Generation settings, retry policy,
The public appendix summarizes the scoring and evaluation boundary; implementation
materials are provided in the release artifact.

\vspace{-0.4em}
\subsection{How Well Do Current Models Solve Visual Problems? (RQ1)}
\vspace{-0.4em}

\paragraph{Overall performance.}
Table~\ref{tab:benchmark_results} reports semantic, relational, and visual-quality scores for required and protected content, together with task completion $R$ and collateral damage $D$. GPT-Image-2 obtains the highest \solvescore at \result{57.0\%} among the eleven systems evaluated here, followed closely by Seedream 5.0 Pro (\result{56.6\%}) and Gemini 3.1 Flash Image (\result{55.5\%}). Commercial image editors substantially outperform the open-source systems, while the image-to-video models remain weaker under the final-frame protocol. Even the strongest model leaves substantial completion and preservation errors.

\vspace{-0.8em}
\paragraph{Completion and preservation.}
Similar aggregate scores can conceal different completion and preservation
profiles. GPT-Image-2 has the highest task-completion score
($R=\result{67.6\%}$), but incurs more collateral damage than Seedream
($D=\result{23.9\%}$ versus $D=\result{17.6\%}$). Seedream therefore reaches
a similar overall score through better preservation rather than higher
completion. The near tie does not imply interchangeable behavior: improving
task completion alone can leave substantial unintended changes unaddressed.
Moreover, required visual quality exceeds required semantic and relational
accuracy for all three leading models in Table~\ref{tab:benchmark_results}.
Visually coherent outputs thus remain an incomplete indicator of task success.
These diagnostics identify which conditions fail; the regime-wise analysis
below examines how these failures relate to the information needed to determine
the transition.

\begin{figure}[t]
\centering
\vspace{-1.5em}
\IfFileExists{figures/fig5.png}{%
    \includegraphics[width=\textwidth]{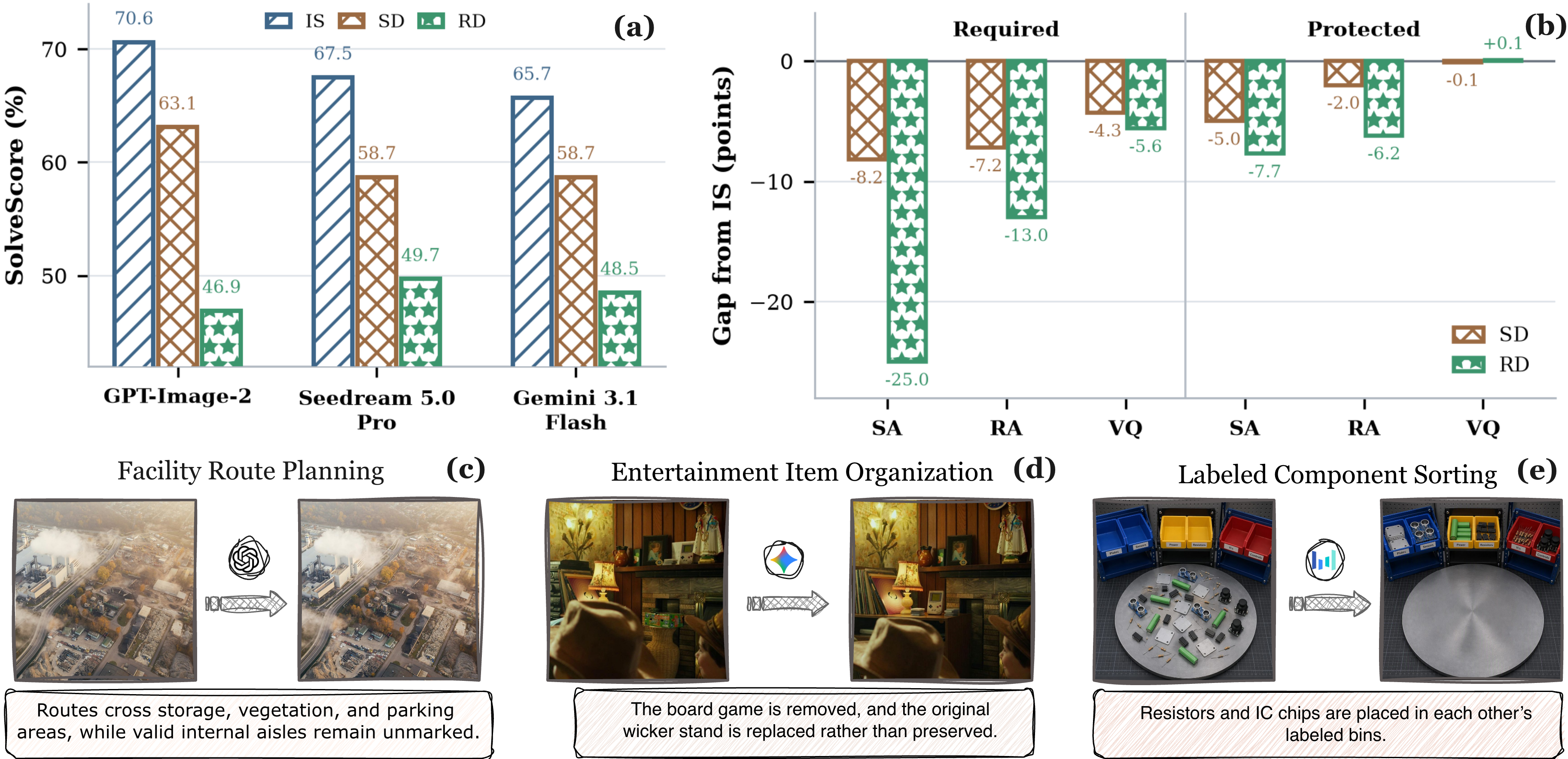}%
}{%
    \fbox{\parbox[c][6.1cm][c]{0.90\textwidth}{\centering
    \textit{Reasoning failures behind visually plausible outputs.}}}%
}
\vspace{-1.5em}
\caption{Reasoning failures behind visually plausible outputs. \textbf{Top left:} \solvescore across IS, SD, and RD for the three leading models. \textbf{Top right:} application-domain-adjusted diagnostic gaps from IS, pooled across models and separated into required and protected conditions. \textbf{Bottom:} representative outputs that remain visually coherent but violate task-specific spatial, semantic, or rule-based requirements.}
\vspace{-1.5em}
\label{fig:rq2_dependency_analysis}
\end{figure}

\vspace{-0.4em}
\subsection{What Do Visually Plausible Failures Reveal? (RQ2)}
\vspace{-0.4em}

\paragraph{Performance declines with information dependence.}
Current models often produce coherent edits that nevertheless fail the task. We compare the three information-dependence regimes, IS, SD, and RD, and separate failures to satisfy required conditions from collateral changes to protected content. For each diagnostic cell, we compare regimes within application domains and aggregate the differences using common domain weights. This adjustment reduces the influence of domain composition, helping distinguish the observed regime gaps from differences in the mixture of task categories; the public appendix documents the regime labels and scoring protocol.

\vspace{-0.8em}
\paragraph{Diagnostic gaps.}
Figure~\ref{fig:rq2_dependency_analysis}(a) shows a consistent gap from IS to RD for three leading models: RD trails IS by \result{17.2 to 23.7 points}, while the SD gaps range from \result{7.0 to 8.8 points}. This ordering is consistent with the additional information required to determine the transition: IS cases can be solved from the grounded request, whereas SD and RD cases require recovery of scene state or an in-image rule. After adjusting for domain composition, Figure~\ref{fig:rq2_dependency_analysis}(b) shows that the largest deficits from IS to RD occur in required semantic (\result{-25.0} points) and relational (\result{-13.0} points) conditions, compared with \result{-5.6} points for required visual quality; protected dimensions show smaller deficits (\result{-7.7}, \result{-6.2}, \result{+0.1} for SA, RA, VQ), with visual-quality preservation near neutral. This asymmetry indicates that the regime gap is concentrated in satisfying the requested change, with a much smaller deterioration in preserving unrelated content. Together with the qualitative examples in Figure~\ref{fig:rq2_dependency_analysis}(c) to (e), this pattern suggests that solution-recovery errors are a major contributor, rather than the failures arising only from visual execution. The regimes describe information dependence, not a prescribed difficulty ranking. This motivates testing explicit transition planning.

\vspace{-0.4em}
\subsection{Does Explicit Transition Determination Improve a Fixed Generator? (RQ3)}
\vspace{-0.4em}

\paragraph{Matched comparison.}
\begin{sloppypar}
We compare Direct, Text-only Rewrite, \genericvisionrewrite, and \solveeditplan
under a matched one-call final-generation budget from the same fixed editor.
This holds the case manifest, output count, and scoring contract constant while
testing whether explicit transition inference helps beyond longer instructions.
The ablations remove contrastive checking or preservation inference; protocol
details are provided in Appendix~\ref{app:solveedit_plan_results}.
\end{sloppypar}

\begin{figure}[!t]
\centering
\vspace{-1.5em}
\IfFileExists{figures/figure6_planning_results.png}{%
    \includegraphics[width=\textwidth]{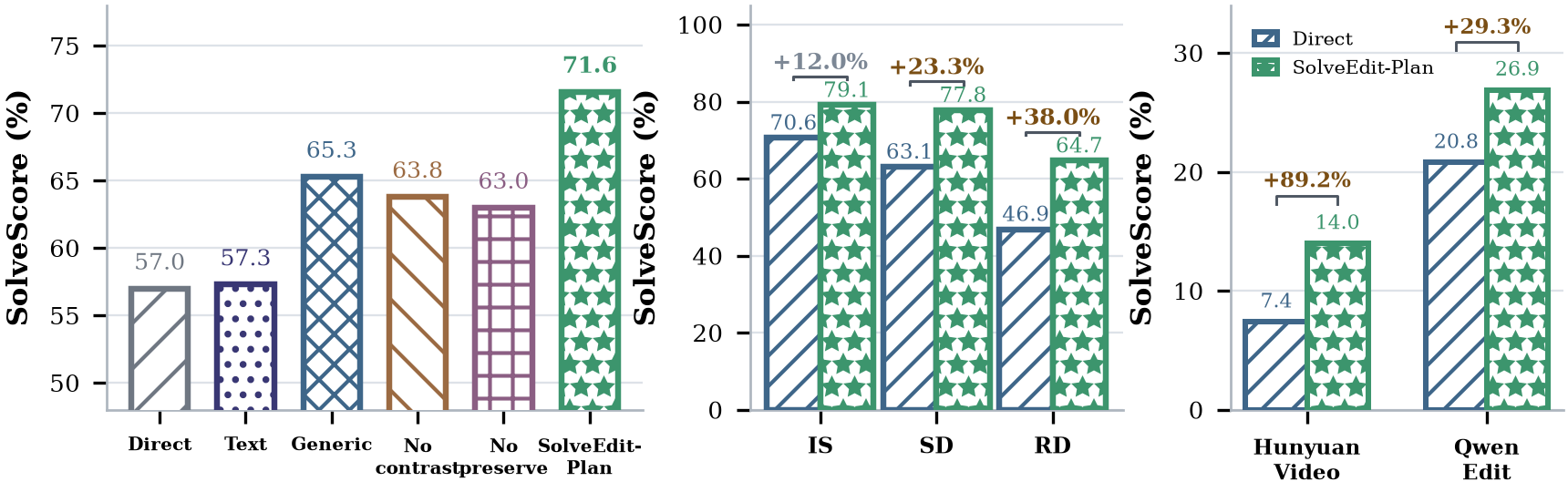}%
}{%
    \fbox{\parbox[c][3.8cm][c]{0.96\textwidth}{\centering
    \textit{Planned Figure 6. Left: GPT-Image-2 controls and ablations. Middle: gains by dependency level. Right: Direct versus \solveeditplan across fixed generators.}}}%
}
\vspace{-2em}
\caption{Effect of explicit transition inference. \textbf{Left:} GPT-Image-2 controls and component ablations. \textbf{Middle:} GPT-Image-2 gains by dependency level. \textbf{Right:} Direct versus \solveeditplan across fixed generators. Every method uses one final generation.}
\vspace{-1.5em}
\label{fig:planning_results}
\end{figure}

\vspace{-0.8em}
\paragraph{Main gains and ablations.}
\solveeditplan improves GPT-Image-2 from \result{57.0} to \result{71.6} \solvescore, outperforming the budget-matched \genericvisionrewrite by \result{6.3} points. The gain transfers to the tested open-source image editor and video generator without changing the parameters of either final generator. Compared with generic rewriting, required completion increases from \result{72.8} to \result{75.8}, while collateral damage decreases from \result{16.3} to \result{8.6}. The larger gains on SD and RD are consistent with the information-dependent failures diagnosed in RQ2; full ablations and regime-wise results are reported in Appendix~\ref{app:results}.
Figure~\ref{fig:planning_results} summarizes the controls, ablations, regime-wise gains, and cross-generator transfer. Additional per-category and per-regime results, generation success, evaluator coverage, $\lambda$ sensitivity, and component ablations are reported in Appendix~\ref{app:results} for completeness.

\vspace{-0.4em}
\subsection{Further Analysis: Cost and Cross-Generator Transfer}
\vspace{-0.4em}

\paragraph{Planning cost.}
The planning intervention adds a measurable test-time cost. Each case uses two
LLM calls, one for Inspect and one for Resolve, with an average of 3{,}207
input tokens and 635 output tokens. This adds computation before the same
single final generation, so the gains over Direct are obtained with additional
planning computation. The stronger control is \genericvisionrewrite:
with the same planning backend and call budget, \solveeditplan achieves a
higher score on GPT-Image-2, supporting the value of structured transition
planning within that budget. Equal call counts do not imply equal token use;
implementation-dependent latency and cost accounting are provided in
Appendix~\ref{app:solveedit_plan_results}.

\vspace{-0.8em}
\paragraph{Transfer across generators.}
The improvement is not limited to GPT-Image-2: under the same case manifest and
transition contracts, \solveeditplan increases \solvescore from 20.8\% to
26.9\% for Qwen-Image-Edit-2509 and from 7.4\% to 14.0\% for HunyuanVideo-1.5.
Both generators improve required completion and reduce collateral damage
(Table~\ref{tab:app_planning_results}), showing that the gains extend to both
parts of the transition contract. This supports the applicability of the
planning procedure through the instruction interface for the tested image and
video generators, despite their different generation architectures and output
modalities. Their low absolute scores nevertheless leave substantial
residual error: these results establish neither reliable task completion nor
whether the remaining failures originate in planning or generation.

\section{Conclusion}
\label{sec:conclusion}
\vspace{-0.4em}

We introduced visual problem solving through generative transformation of an existing visual scene. In this setting, a model must determine a valid transition from the request and visual evidence, realize it in the output, and preserve unrelated content. \solveedit provides 2,728 cases organized by the information that determines the required change, while atomic transition contracts and \solvescore measure completion and preservation without requiring a single reference output. Across the evaluated generators, the strongest model achieves only 57.0\% \solvescore, with larger deficits on scene- and rule-dependent transitions than on instruction-specified cases. \solveeditplan provides a targeted test of this diagnosis: explicitly recovering the transition before one call to an unchanged generator improves GPT-Image-2 from 57.0\% to 71.6\% under a matched final-generation budget and transfers to the tested open-source image editor and video generator. Together, these results establish a controlled benchmark for studying transition determination, visual execution, and preservation in generative visual problem solving.

\bibliography{references}
\bibliographystyle{plainnat}

\appendix
\clearpage
\setlength{\parskip}{4pt}
\raggedbottom

\section{Benchmark Annotation and Evaluation}
\label{app:dataset}

\subsection{Benchmark composition}

\solveedit contains 2,728 cases across 10 application domains and 54 canonical
subdomains. Application domain, solution-dependency regime, and image
provenance are separate annotations. The final source composition is as follows:

\begin{table}[H]
\centering
\scriptsize
\caption{Final source composition of the \solveedit benchmark.}
\label{tab:app_source_composition}
\begin{tabular}{lr}
\toprule
Source category & Cases \\
\midrule
AI-generated inputs & 1,375 \\
Real photographs & 646 \\
Animation and game captures & 377 \\
Film and cinematic frames & 330 \\
\midrule
Total & 2,728 \\
\bottomrule
\end{tabular}
\end{table}

These categories describe image provenance rather than visual content or task
regime. The release records source identifiers, image hashes, provenance notes,
and applicable usage information. Cases are retained only when the intended
change is visually checkable, the evidence needed to determine it is present,
correctness can be expressed without exact reference matching, and required
changes can be separated from protected content. Ambiguous or inaccessible
cases are revised or withheld.

\subsection{Dependency-regime annotation}
\label{app:annotation}

Annotators first bind request terms to visible entities and regions. They then
apply the following decision sequence:

\begin{enumerate}
    \item Assign IS when the grounded request and relevant background knowledge
    determine the semantic action and valid outcome.
    \item Otherwise, assign SD when observable current-scene facts determine the
    missing target, action, destination, route, or final state.
    \item Assign RD when a case-local rule, legend, reference relation,
    capacity, pattern, or compatibility condition must additionally be read or
    induced from the image.
\end{enumerate}

The label records the minimum case-specific information needed to determine the
transition, not task difficulty. Ordinary grounding remains IS. Background
knowledge alone does not make a case RD unless a case-local rule is shown or
instantiated in the image. Each assignment records the unresolved variable and
the visual evidence used to resolve it.

\begin{tcolorbox}[breakable,colback=gray!6,colframe=black!55,boxrule=0.5pt]
\small
\textbf{Boundary examples.} ``Move the red marker to the left side of the
board'' is IS after grounding. ``Move the marker to the available slot'' is SD
when occupancy determines the destination. ``Arrange the markers according to
the legend shown in the image'' is RD when the visual legend determines the
valid arrangement.
\end{tcolorbox}

\subsection{Evaluation information boundary}

The editor receives only the source image and intent-level request. The
evaluator receives the source image, request, model output, and the full hidden
transition contract. It cannot revise the request, use a reference output to
invent a target, or use the model identity as evidence. When visible evidence
is insufficient, the evaluator returns \texttt{abstain}; the abstention remains
in the audit record and reduces evidence coverage.

\section{Scoring and Evaluation Details}
\label{app:contracts}

Each case stores required postconditions and independent protected conditions.
Every atomic criterion has an observable question, pass/partial/fail
conditions, an evidence scope, a role, and one diagnostic property. Required
criteria describe completion; protected criteria describe preservation.

\subsection{Scoring protocol}

The evaluator first applies the quality gate to reject missing, blank, severely
corrupted, unrelated, or unusable outputs. It then assigns each applicable atom
one of \{pass, partial, fail, abstain\}. The deterministic scorer maps pass,
partial, and fail to $a_q\in\{1,0.5,0\}$, respectively, and computes
\begin{equation}
    R=\sum_{q\in\mathcal{C}^{\mathrm{req}}}w_q a_q,
    \qquad
    D=\sum_{q\in\mathcal{C}^{\mathrm{pro}}}u_q(1-a_q),
\end{equation}
\begin{equation}
    \solvescore_{\lambda}=G_{\mathrm{quality}}
    \max(0,R-\lambda D).
\end{equation}
The primary protocol fixes $\lambda=0.5$. Required and protected weights are
normalized separately, with equal mass for applicable SA, RA, and VQ properties
within each role. Abstained criteria retain their original weights but
contribute zero to the corresponding sum. No weights are redistributed.

If all required criteria abstain, $R=0$; if all protected criteria abstain,
$D=0$, which indicates no established violation rather than verified
preservation. Output-induced deletion, occlusion, or distortion that violates
a criterion is scored as fail or partial according to its rubric. Evidence
coverage is the non-abstained criterion weight divided by the total criterion
weight across both roles.

\paragraph{Edge-case handling.}
\textbf{Quality-gate failure:} a missing, blank, severely corrupted, unrelated,
or unusable output receives \solvescore~0. \textbf{Generation failure:} a model
that produces no usable image after at most two retries remains in the full
denominator and receives score 0. \textbf{All-criteria abstain:} a case for
which every applicable atom abstains is scored as 0 and retained in the audit
record. \textbf{Partial coverage:} coverage differences are reflected in the
score rather than hidden by subset averaging.

\subsection{VLM evaluation and specialized tools}

The main evaluation uses GPT-5.6 Sol at temperature~0. Fixed checker
assignments are shared across evaluated models for criteria where a specialized
tool provides useful evidence. SAM and YOLO provide segmentation and detection
evidence for assigned object and region conditions. Tool predictions support
criterion assessment but are not assumed to be error-free. For assigned
criteria, checker and VLM verdicts are reconciled by one additional VLM call
when they disagree; unassigned criteria use the VLM verdict directly.

\begin{table}[H]
\centering
\scriptsize
\caption{Evidence families used for atomic criteria.}
\label{tab:app_metric_suite}
\begin{tabular}{p{0.23\linewidth}p{0.30\linewidth}p{0.35\linewidth}}
\toprule
Evidence family & Typical criteria & Evidence produced \\
\midrule
OCR and parsers & Text, symbols, digits, labels, state updates & Recognized string/state and normalized comparison \\
Grounding and segmentation & Entity presence, count, target zones & Instances, boxes/masks, count and containment \\
Geometry and topology & Position, alignment, routes, occupancy, assembly & Distances, overlaps, adjacency and graph relations \\
Preservation and similarity & Protected objects, layout, background, quality & Region-level change and similarity signals \\
VLM contract audit & Scene interpretation and criterion assessment & Pass/partial/fail/abstain verdict and visible evidence \\
\bottomrule
\end{tabular}
\end{table}

\subsection{Diagnostic dimensions}

Semantic Accuracy (SA) covers entities, identity, count, attributes, text,
symbols, and intrinsic state. Relational Accuracy (RA) covers position, order,
containment, alignment, connectivity, topology, and occupancy. Visual Quality
(VQ) covers rendering defects such as residue, blur, malformed local geometry,
and illegible content. The canonical contract inventory contains 29,460 atomic
criteria: 14,831 required and 14,629 protected; 14,322 are SA, 10,520 are RA,
and 4,618 are VQ. The mean and median numbers of criteria per case are 10.8
and 10, respectively, and every case has at least one protected criterion.

\subsection{Worked score calculation}

Suppose a case has four equally weighted required atoms with verdicts pass,
pass, partial, and fail, and two equally weighted protected atoms with verdicts
pass and partial. Then $R=0.625$, $D=0.25$, and the primary score is
$\max(0,0.625-0.5\times0.25)=0.50$, assuming the quality gate passes.

\section{Planning Details}
\label{app:solveedit_plan_results}

\solveeditplan receives only the source image and task request. It cannot access
the evaluation contract, reference output, authored regions, manual facts,
dependency label, or evaluator response. Inspect identifies unresolved
variables and crop queries; Resolve compares candidate transitions and compiles
one instruction for one final generator call. Direct, Text-only Rewrite,
\genericvisionrewrite, ablations, and \solveeditplan use the same final
generation budget in the matched comparison.

\begin{table}[H]
\centering
\scriptsize
\caption{Matched one-generation evaluation of \solveeditplan and controls (\%).}
\label{tab:app_planning_results}
\begin{tabular}{llcccc}
\toprule
Generator & Instruction method & $R\uparrow$ & $D\downarrow$ & \solvescore$\uparrow$ & $\Delta$ \\
\midrule
\multirow{2}{*}{GPT-Image-2} & Direct & 67.6 & 23.9 & 57.0 & -- \\
& \textbf{\solveeditplan} & 75.8 & 8.6 & 71.6 & +14.6 \\
\midrule
\multirow{2}{*}{Qwen-Image-Edit-2509} & Direct & 32.7 & 47.5 & 20.8 & -- \\
& \textbf{\solveeditplan} & 40.5 & 38.5 & 26.9 & +6.1 \\
\midrule
\multirow{2}{*}{HunyuanVideo-1.5} & Direct & 17.3 & 55.5 & 7.4 & -- \\
& \textbf{\solveeditplan} & 23.7 & 39.7 & 14.0 & +6.6 \\
\midrule
\multicolumn{6}{c}{\textit{GPT-Image-2 controls}} \\
& Text-only Rewrite & 67.9 & 23.2 & 57.3 & +0.3 \\
& \genericvisionrewrite & 72.8 & 16.3 & 65.3 & +8.3 \\
& \textbf{\solveeditplan} & 75.8 & 8.6 & 71.6 & +14.6 \\
\bottomrule
\end{tabular}
\end{table}

\paragraph{Planner configuration.}
The released implementation uses GPT-5.6 Sol (API identifier
\texttt{gpt-5.6-sol}) with temperature~0. Inspect receives the source image;
Resolve receives it together with at most six crops requested by Inspect.
\genericvisionrewrite uses the same model and call budget with no additional
crops. \solveeditplan calls Inspect and Resolve once each; both methods
use one final generation call.

\label{app:results}
All main results use the same 2,728-case manifest and deterministic scorer.
Generation failures remain in the denominator, and video outputs are scored
from their terminal frames. Full prompts, schemas, and intermediate records
are provided with the release artifact.

\section{Representative Cases}
\label{app:case_cards}

The following cases illustrate the three information-dependence regimes and
the distinction between completing the requested transition and preserving
unrelated content. They are representative examples rather than an exhaustive
gallery; the project page provides a browsable presentation.

\paragraph{Case 1 (IS): Selective cleanup.}
The request explicitly identifies the fallen debris to remove from a carpet,
while intact objects and furniture must remain unchanged. The scene grounds
the referenced categories but does not determine the intended action.

\begin{figure}[H]
\centering
\includegraphics[width=0.47\linewidth]{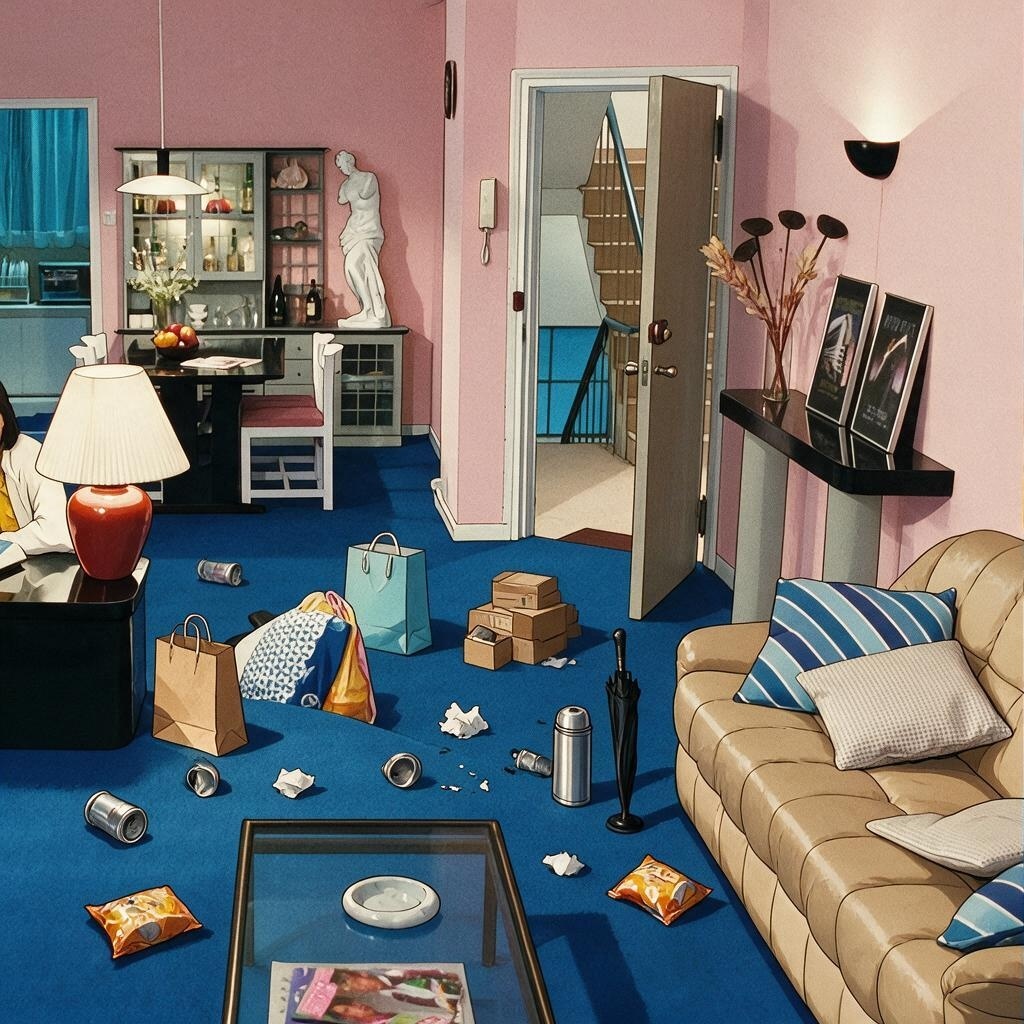}
\hfill
\includegraphics[width=0.47\linewidth]{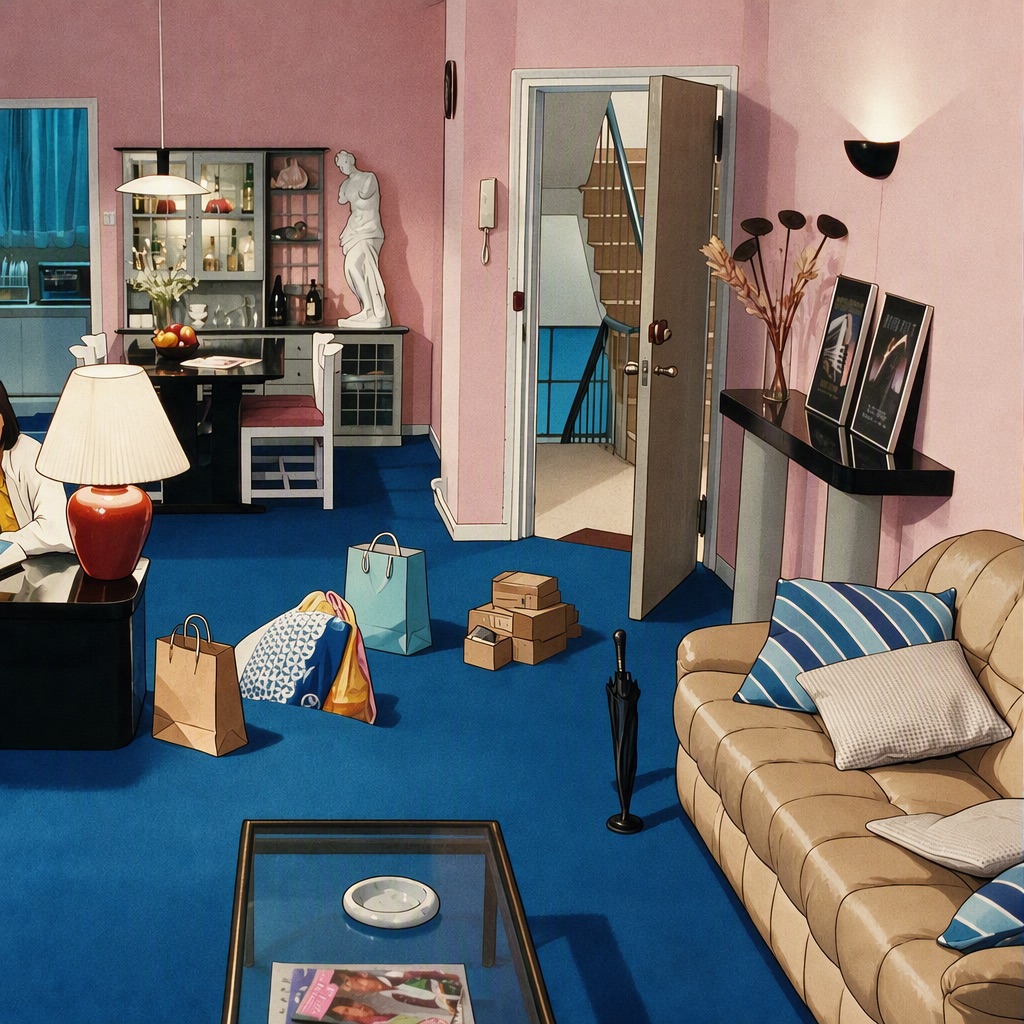}
\caption{IS case: selective cleanup. Left: input; right: GPT-Image-2 output.}
\label{fig:app_case_is}
\end{figure}

\paragraph{Case 2 (SD): Road bridge repair.}
The request asks for missing roads to be repaired, but the current scene fixes
which bridge segment belongs over the river gap through path endpoints and
matching texture. Choosing the wrong segment produces a plausible image that
does not satisfy the transition contract.

\begin{figure}[H]
\centering
\includegraphics[width=0.47\linewidth]{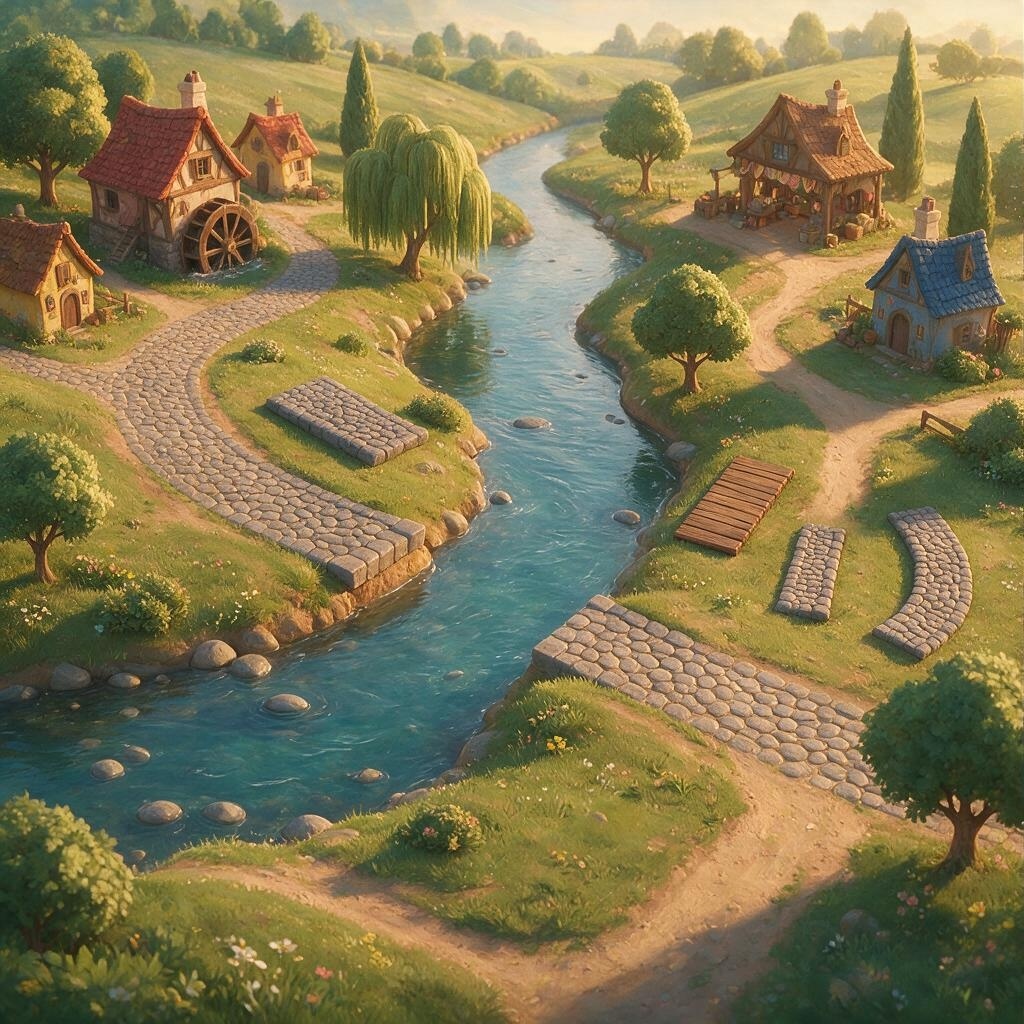}
\hfill
\includegraphics[width=0.47\linewidth]{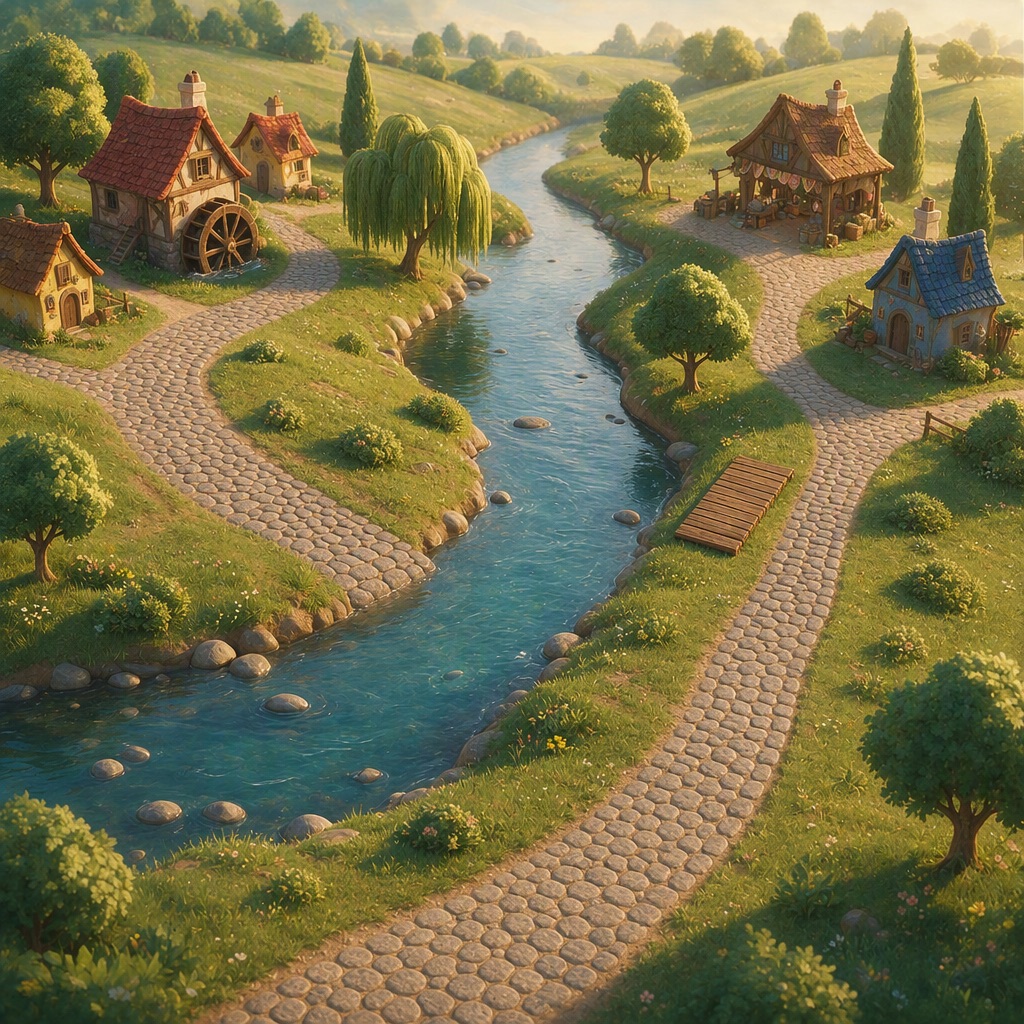}
\caption{SD case: road bridge repair. Left: input; right: GPT-Image-2 output.}
\label{fig:app_case_sd}
\end{figure}

\paragraph{Case 3 (RD): Storyboard connection repair.}
The in-image grid and sequence rules determine where cue cards and tempo
markers belong. The edit must repair broken connections while preserving panel
content and readable text.

\begin{figure}[H]
\centering
\includegraphics[width=0.47\linewidth]{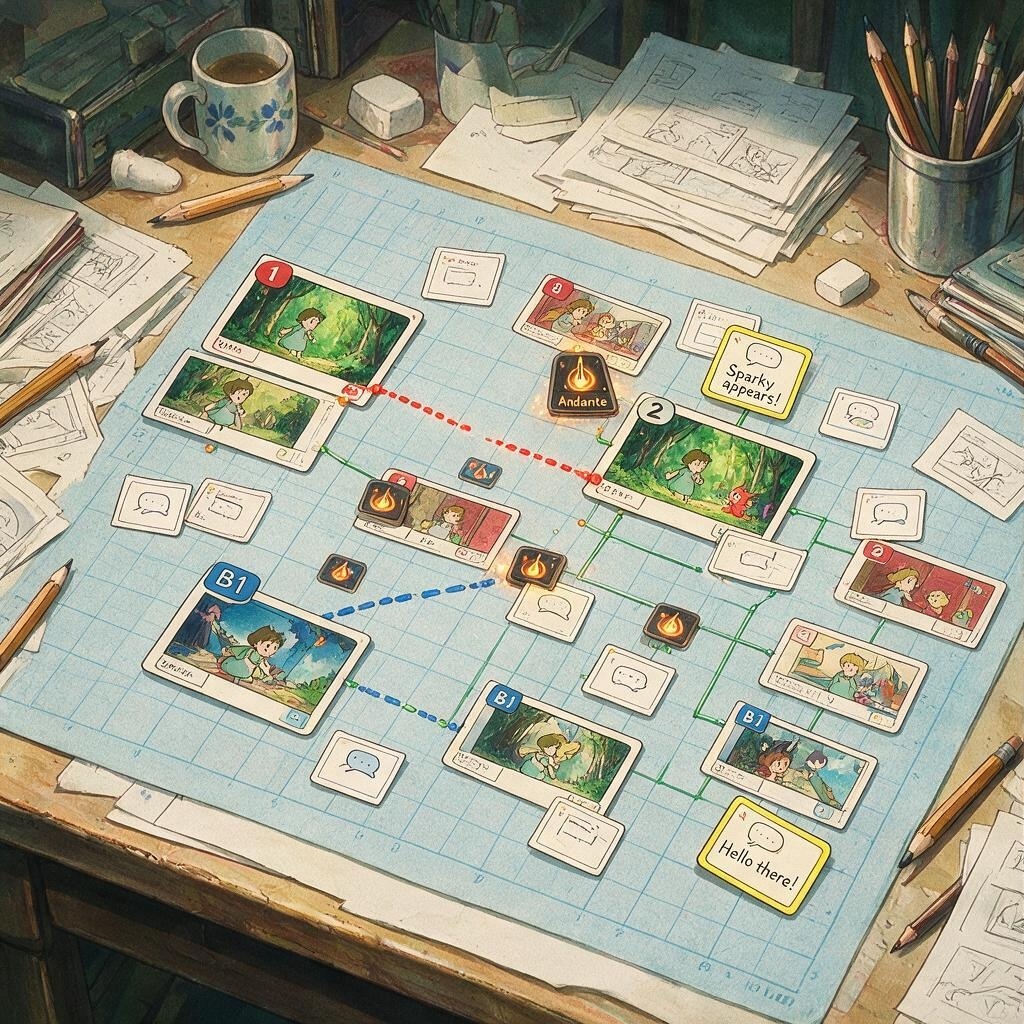}
\hfill
\includegraphics[width=0.47\linewidth]{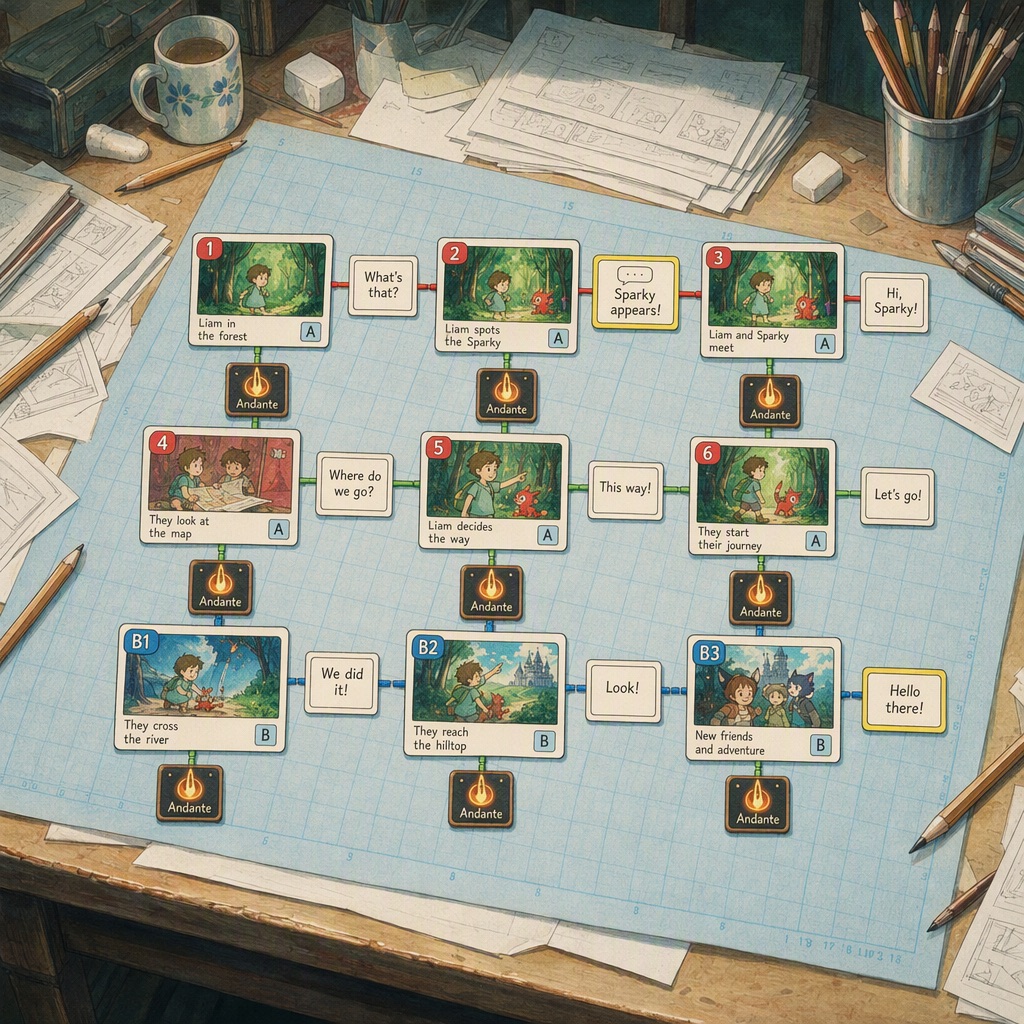}
\caption{RD case: storyboard connection repair. Left: input; right: GPT-Image-2 output.}
\label{fig:app_case_rd}
\end{figure}

\end{document}